\documentclass[conference]{IEEEtran}
\usepackage[cmex10]{amsmath}
\usepackage{tabularx}
\usepackage[table]{xcolor}
\usepackage{cleveref}
\usepackage{graphicx}
\usepackage{multirow}
\usepackage{amssymb}
\usepackage{subfigure}
\usepackage{upquote}
\usepackage{soul}
\graphicspath{{media/}}
\usepackage{ntheorem}

\usepackage{cite}
\usepackage{url}
\RequirePackage{subfigure}
\usepackage{pdfpages}
\usepackage{float}
\usepackage{subfig}
\usepackage{array}
\pdfoutput=1
\begin{document}

\title{Human Brains Can't Detect Fake News: \\A Neuro-Cognitive Study of Textual Disinformation Susceptibility}

\author{\IEEEauthorblockN{Cagri Arisoy}
	\IEEEauthorblockA{Texas A\&M University\\
		cagri.arisoy@tamu.edu}	
	\and
	\IEEEauthorblockN{Anuradha Mandal}
	\IEEEauthorblockA{The University of Alabama at Birmingham\\
		anuradha@uab.edu}
	\and
	\IEEEauthorblockN{Nitesh Saxena}
	\IEEEauthorblockA{Texas A\&M University\\
		nsaxena@tamu.edu}
	}

\maketitle

\begin{abstract}

The spread of digital disinformation (aka ``fake news'') is arguably one of the
	most significant threats on the Internet today which can cause
	individual and societal harm of large scales. The susceptibility to
	fake news attacks hinges on whether or not Internet users perceive a
	fake news article/snippet to be legitimate (real) after reading it.
	In this paper, we attempt to garner an in-depth
	understanding of users' susceptibility to \textit{text-centric} fake news attacks via a
	neuro-cognitive methodology (thus corroborating as well as extending the
	traditional behavioral-only approach in significant ways).
	In particular, we investigate the neural underpinnings relevant to fake
	vs. real news through EEG, a well-established brainimaging technique.
	We design and run an EEG experiment with human users to pursue a
	thorough investigation of users' perception and cognitive processing of fake vs. real
	news. We analyze the neural activity associated with the fake vs. real
	news detection task for different categories of news articles. 


Our results show that there may be no statistically significant or
	automatically inferable differences in the way the human brain
	processes the fake vs. real news, while marked differences are observed
	when people are subject to (real or fake) news vs. resting state and
	even between some different categories of fake news.  This
	neuro-cognitive finding may help to justify users' susceptibility to
	fake news attacks, as also confirmed from the behavioral analysis. In
	other words, the fake news articles may seem almost indistinguishable
	from the real news articles in both behavioral and neural domains. 
	Our work serves to dissect the fundamental neural phenomena underlying fake
	news attacks and explains users' susceptibility to these attacks
	through the limits of human biology.  We believe that this could be a
	notable insight for the researchers and practitioners suggesting that
	the human detection of fake news might be ineffective, which may also
	have an adverse impact on the design of automated detection approaches
	that crucially rely upon human labeling of text articles for building training models.
	
	

 \end{abstract}

\section{Introduction}
\label{sec:intro}

The spread of social disinformation on the Internet (informally often referred
to as ``fake news'') is an emerging class of immensely powerful threats, especially given
the widespread deployment of social media, seen in a variety of contexts that
can severely harm everyday Internet users and the society in many different
ways at a global scale. Fake news attacks are used, for instance, to earn money
from advertisements, \cite{news-advertisement}, defame an agency, entity, or person, impact people's behavior or even influence the results of
elections as seen in the last mainstream elections in the US, India and Brazil \cite{election1,election2,election3}.
The susceptibility to such fake news attacks depends on whether users consider
a fake news article/snippet to be legitimate or real. Unfortunately, behavioral
studies demonstrate that the general user population may often believe fake news
to be real news and indeed be fallible to fake news attacks \cite{Vosoughi1146,michaelbarthel}.

Due to the prevalence and rapid emergence of the fake news threat in the wild,
it is paramount to understand users' intrinsic psychological behavior that
controls their processing of (fake and real) news articles and their potential
susceptibility to fake news attacks.  In this paper, we employ the brain-imaging
methodology adopted in a recently introduced line of research
(e.g., \cite{andersen-fmri,neupaneEEG,neupane2014neural}) to closely assess the user
behavior in the specific context of disinformation susceptibility.
Specifically, we study users' neural processes (in addition to their behavioral
performance) to understand and leverage the {neural} mechanics when
users are subjected to fake news attacks using a state-of-the-art,
well-established brain-imaging technique called \textit{electroencephalogram
(EEG)}.  

Our primary goal is to study the {neural underpinnings} of fake news detection,
and analyze differences (if any) in neural activities when users process (read
and decide between) fake and real news. We examine how the information present
in the neural signals may be used to explain users' susceptibility to fake news
attacks. \textbf{We focus on \textit{text-centric} news articles where only the most
fundamental information inherent to the article, i.e., the title and textual
content, is shown to the user (and no auxiliary information such as images, the
URL, hosting website or the advertisements embedded within the site)}.  Such
text-centric fake news is often spread via social media posts or text
messaging platforms.

Some of the prior studies
{\cite{huang2011human,neupaneEEG,neupanefNIRS,neupane2014neural}} have shown
that {neural differences are present when users are subject to different types
of real vs. fake artifacts, even though users themselves may not be able to
tell the two apart behaviorally}. 
Neupane et al.  \cite{neupaneEEG,neupane2014neural,neupanefNIRS} noted
differences in neural activities 
when users were processing real and fake websites in the context of phishing
attacks.  Similarly, Huang et al.  \cite{huang2011human} reported differences
{in neural activation} 
when users were asked to assess real and fake Rembrandt paintings. 
%
%
In light of these prior results, we launched our study to also test the
{hypothesis that the human brain might be activated differently when users are
subject to fake and real news}.

The implications of the neural activity differences, if present, when
processing real vs. fake news, can be far-reaching as these differences could
be automatically mined and analyzed in real-time and the user under the fake
news attack could be informed as to the presence/absence of the attack, even
though the user may have himself failed to detect the attack (behaviorally). 
Neupane et al. \cite{neupanefNIRS} suggested such a scheme for detecting phishing attacks merely
based on neural activities.
{On the other hand, it is crucial to note that in case the differences do not exist, 
it may underline a fundamental vulnerability of human biology which may prevent users 
from telling the fake artifact from the real artifact.}
Our study is designed to
investigate the same aspect, but in the independent application domain of fake
news attacks and disinformation susceptibility.  In this domain, the user is
supposed to be detecting fake news based on the \textit{title/content of the news}, not based
on the side information such as the URL hosting the news since the fake news can
often appear on legitimate websites. This contrasts with the domain of phishing
attacks where URLs are the most important indicators of the presence of the
attack.   
\noindent \textbf{{Our Contributions:}} \label{Our Contributions and
Results} 
We design and conduct an EEG study to pursue a thorough investigation
of users' perception and mental processing of fake and real news. Our study asks the participants to
identify real vs. fake news articles, drawn from the study of Zannettou et al.
\cite{ZannettouCCKLSS17}, presented in a randomized order solely based on the title/content of the articles.  We provide a comprehensive analysis of
the collected EEG data set including the EEG frequency bands (such as alpha, beta, and gamma), the 
aggregated EEG metrics (attention and meditation) and the raw EEG signals,  
as well as the behavioral task performance data
set, across different categories of news articles (political, daily, and miscellaneous news). 

Unlike prior studies on real-fake websites and paintings quoted above,
we do not observe differences in the way the brains process fake news vs. real
news, when subject to fake news attacks, although marked differences are seen
between neural activity corresponding to (real or fake) news vs. resting state
(participants not doing any mental activity) and between some of the different
categories of news articles (such as daily fake news vs. other miscellaneous
fake news).  That is, the fake news seems nearly indistinguishable from the real
news with respect to the neuro-cognitive perspective.  This rather negative/null result  may serve
well to explain users' susceptibility to such attacks as also reflected in our
task performance results (similar to those reported in 
\cite{Vosoughi1146,michaelbarthel}). We further tested several machine learning algorithms based on statistical features
and even they failed at distinguishing between fake and real news with a probability
significantly better than 50\% (equivalent to random guessing).

Since this very likely indistinguishability of real vs. fake news lies at the
core of human biology, we conjecture that the problem is very critical, as the human
detection of fake news may not improve over time even with human evolution, especially
because malicious actors may continue to come up with more advanced and surreptitious ways to 
design fake news. 
Also, our study participants are
mostly young and educated individuals and with no reported visual disabilities, while older/less educated
population samples and/or those having visual disabilities may be more vulnerable
to fake news attacks \cite{dubno1984effects}.

{We cautiously \textit{do not} claim that the acceptance of the null hypothesis in our work necessarily means that the neural differences
		between real and fake news are definitively absent, i.e., further studies might need to be conducted using
		other brain-imaging techniques and other wider samples of users.
		However, our work does cast a serious doubt
		regarding the presence of such differences 
which is also aligned well with our behavioral
		results, thereby explaining the high potential of user-centered disinformation susceptibility.  

The presence of neural activity differences between some types of fake news
seen in our study could also have interesting implications. For instance, it
could be used by malicious players to design targeted fake news attacks.


		\smallskip \noindent \textbf{Broader Significance, and Negative
Implications to Automated Detection Techniques:} We believe that our work
helps to advance the science of human-centered disinformation susceptibility,
in many unique ways. It also serves to reveal the fundamental neural basis
underlying fake news attacks, and highlights users' susceptibility to these
attacks from both neural as well as behavioral domains.  

In light of our results, perhaps the best way to protect users from
such attacks would be by making them more aware of the threat, and possibly by
developing technical solutions to assist the users. 
The security and disinformation community has certainly been working towards
urgently developing automated fake news detection techniques which could aid the end
users against fake news-based scams, whenever possible. However, there is a critical cyclic dependency here. Since these automated techniques themselves
rely upon human-based or crowd-sourced labeling of fake vs. real news to train (and re-train) the underlying classification models (e.g., \cite{oshikawa2018survey,8966734,wang2017liar, HadeerA2017, Granik, FIGUEIRA2017817}), our work 
raises a question on the robustness of such models when used in the real-world.   
		 
		
		
		

		\smallskip \noindent \textbf{Why is this a Computer Security Study?} Although our
		work is informed by neuroscience, it is deeply rooted in computer security and provides
		valuable implications for the security community. First and foremost, disinformation and fake news is a security threat that leads to a large variety of financial and political scams, and social engineering trickeries against casual Internet users\footnote{As a colloquial example of security relevance, disinformation was a topical area of a keynote speech at the 2019 NSF SaTC PI meeting.}. Said differently, detecting fake news is a user-centered security task, similar to detecting phishing or other similar attacks.
		{Second, we conduct a neuroimaging-based user study and show why malicious actors might be successful at fake news attacks.} Many similar security studies focusing on human
		neurobiology have been published as an ongoing line of research in mainstream
		security/HCI venues,
		{e.g.,~\cite{andersen-fmri,neupaneEEG,neupanefNIRS,neupane2014neural}}. How users
		perform at crucial security tasks from a neurological standpoint is therefore
		of great interest to the security community.
		
		This research line followed through in our work provides novel security insights and lessons
		that are \textit{not feasible} to elicit via behavioral studies {alone}.
		For example, prior studies {\cite{neupaneEEG,neupanefNIRS,neupane2014neural}}
		showed that security (phishing) attacks can be detected based on neural cues,
		although users may themselves not be able to detect these attacks. Along
		this dimension, our work conducted an EEG study to dissect users' behavior under
		fake news attacks, a rather understudied attack vector. Our results show
		that even brain responses cannot be used to detect such attacks, which serve to
		explain why users are so susceptible to these attacks.
		

\section{Background \& Prior Work}
\label{sec:back}

\subsection{EEG Overview}
\label{eegoverview}
{
Electroencephalograph (EEG) is a technique to measure the electrical activity that is generated by the brain. EEG’s significant sequential solution is used for measuring the time progression in brain activation among the different regions of the brain. For adults, the amplitude of the signal is between 1uV to 100uV and the range is approximately 10mV to 20mV when it is measured with subdural electrodes \cite{neuroskybrainwave}. Today, EEG is used in several applications such as diagnoses of sleep orders, anesthesia, and tumor. Neuroscience, cognitive science, and many other academic fields routinely benefit from brain signal measurements using EEG.
}
\subsection{EEG Device and Neural Metrics Used}
{
In our study, we used a commercial EEG headset called MindWave Mobile 1 manufactured by Neurosky \cite{neuroskymindwave} to collect brain wave signals as people are subject to fake vs. real news. This headset has an adjustable headband, sensor tip/arm, and ear clip. It connects to a computer using Bluetooth. This device reads the raw brain wave signals and reports frequency bands listed in Table \ref{tab:frequencybands}.

In general, an EEG signal is defined concerning the frequency band of the signal. Some examples of the EEG frequency bands defined by Neurosky are Delta, Theta, Alpha, Beta, and Gamma. Table \ref{tab:frequencybands} drawn from \cite{neuroskybrainwave} outlines these bands and the associated frequency ranges and mental states they represent. As we can see, each of the frequency bands helps to characterize different behavioral activities like movements, thinking, and decision making. While Delta is related to deep sleep levels, Theta is associated with imagination. Alpha and Beta waves, on the other hand, imply relaxation and awareness, respectively. Finally, Gamma is implicated in motor functions and mental activities. To read and obtain the values of the Attention and Meditation metrics, we used \textit {ThinkGear} \cite{thinkgearsocket}, which provides us with an ability to connect to the device. Here, Attention captures the user’s focus and attentiveness or engagement level (related to visual processing and decision making), and Meditation represents the users' calmness or relaxation level (related to relaxation) \cite{neuroskymindwave}. Both have a value range from 0 to 100. If we summarize the range of these metrics: 0 to 20 means strongly lowered, 20 to 40 is the reduced level and 40 to 60 is neutral level, while 60 to 80 is slightly elevated and 80 to 100 is strongly indicative levels \cite{neuroskymindwave}.

\begin{table}[t]
	\centering
\scriptsize
\caption{EEG frequency bands and related brain states \cite{neuroskybrainwave}}
	\begin{tabular}{|l|l|p{4.5cm}|}
		\hline
		\textbf{\begin{tabular}[c]{@{}l@{}}Frequency\\ Band\end{tabular}} & \textbf{\begin{tabular}[c]{@{}l@{}}Frequency\\ range\end{tabular}} & \textbf{Mental states and conditions} \\ \hline \hline
		\textbf{Delta} & 0.1Hz to 3Hz & \begin{tabular}[c]{@{}l@{}}Deep, dreamless, sleep, non-\\ REM sleep, unconscious\end{tabular} \\ \hline
		\textbf{Theta} & 4Hz to 7Hz & Intuitive, creative, recall, fantasy, imaginary, dream \\ \hline
		\textbf{Alpha} & 8Hz to 12Hz & Relaxed but not drowsy, tranquil, conscious \\ \hline
		\textbf{Low Beta} & 12Hz to 15Hz & \begin{tabular}[c]{@{}l@{}}Formerly SMR, relaxed yet\\ focused, integrated\end{tabular} \\ \hline
		\textbf{\begin{tabular}[c]{@{}l@{}}Midrange\\ Beta\end{tabular}} & 16Hz to 20Hz & \begin{tabular}[c]{@{}l@{}}Thinking, aware of self \&\\ surroundings\end{tabular} \\ \hline
		\textbf{High Beta} & 21Hz to 30Hz & Alertness, agitation \\ \hline
		\textbf{Gamma} & 21Hz to 30Hz & \begin{tabular}[c]{@{}l@{}}Motor Functions, higher\\ mental activity\end{tabular} \\ \hline
	\end{tabular}
	\vspace{-2mm}
	\label{tab:frequencybands}
\vspace{-1mm}
\end{table}

}

\subsection{Related Work}
\label{Related-Work}
{
Our study centers on user-centered fake vs real news detection, i.e., the users’ ability to determine whether a given news item is real or fake. Researchers have conducted different studies in this area. Most closely relevant is the study reported by Chan et al. \cite{manpuisally} which is a meta-analysis of the factors underlying effective messages to counter attitudes and beliefs based on disinformation. Barthel et al. \cite{michaelbarthel} conducted a study with the findings that most Americans suspect that made-up news is having an impact and has left them confused about basic facts. The study reported almost a quarter of participants stated that they have posted a made-up news story, intentionally or not. Moreover, Americans see fake news as causing a great deal of confusion in general. However, most of them are confident in their ability to identify when news is false. Vosoughi et al. \cite{Vosoughi1146} conducted a study on the spread of true and false news online. They found that falsehood diffused significantly farther, faster, deeper, and more broadly than the truth in all categories of information. These effects were more pronounced for political news than for fake news about terrorism, natural disasters, science, urban legends, or financial information. It was found that fake news was more novel than real news, which suggests that people were more likely to share novel information.
 
Using the cognitive science approach, Pennycook et al. \cite{pennycook2019} investigated whether cognitive factors motivate belief in or rejection of fake news. They conducted two studies in the paper, utilizing the Cognitive Reflection Test (CRT) as a measure of the proclivity to engage in analytical reasoning. Their results suggest that analytic thinking plays an important role in people’s self-inoculation against political disinformation. The reason people fall for fake news is they do not think. Hoang et al. \cite{hoang2012} investigated user vulnerability as a behavioral component in predicting viral diffusion of general information rather than fake news. Wagner et al. \cite{wagner2012} explored user susceptibility to fake news generated by bot activities and limit their definition of ‘susceptible users’ as the ones that interacted at least once with a social bot. Shen et al. \cite{shen2019} focus on several degrees of susceptibility to fake news by applying machine learning methods.

A related problem is fake vs. real website detection under a phishing attack. Several studies have been conducted in this problem space. Dhamija et al. \cite{Dhamija:2006} reported a study in which participants were asked to identify real and fake websites in the context of phishing attacks. They claimed that the participants chose wrong answers almost 40\% of the time. In line with this, Neupane et al. \cite{neupaneEEG,neupanefNIRS,neupane2014neural} conducted an fMRI, EEG, and fNIRS studies of users’ mental processing of real and fake websites. Their results show that there is a significant increase in several areas of the brain when people view fake websites vs. real websites even though users’ accuracy in identifying the legitimacy of the site was close to 50\%. The study of \cite{voice-ndss} investigated users' processing of real vs. fake voices in the context of voice impersonation attacks, with similar findings as to the above studies.

\begin{table}[ht!]	
	\centering
	\footnotesize
	\caption{An Outlook of Real-Fake Studies in Related Works~vs. Our Work}	
	\begin{tabular}{|p{3cm}|p{2cm}|p{2.2cm}|}
		\hline
		{\bf Artifact Type} & {\bf Neural Activity Diff.} & {\bf Behavioral Response Diff.} \\ \hline \hline
		Websites under phishing & Present            & Nearly absent                 \\ \hline
		Paintings              & Present            & Nearly absent                 \\ \hline
		News (Our Work)                  & Absent             & Absent                 \\ \hline
	\end{tabular}
	
	\label{my-label}
\end{table}
	
Stanford History Education Group conducted a study \cite{sheg} in which the participants (school/college students) were asked to find the differences in some media contents, e.g., news and ads, identify real-fake Facebook accounts, test the reliability of Facebook posts (whether a web site is trusted). Our study, in contrast, investigates real vs. fake news detection in behavioral and neural domains.
	
Different from many of the above studies, our work aims at eliciting a deeper understanding of users’ susceptibility to fake news by focusing on not only people’s behavioral responses as to whether the news is real or fake but also their neurological responses captured via EEG during the decision-making process. Moreover, following the results of prior studies \cite{neupaneEEG,neupanefNIRS,neupane2014neural}, we wanted to test whether people’s brain activities differ when subject to fake vs. real news, even if they may not be able to identify the differences behaviorally. So, we have developed the EEG headset configuration to test ``fake news detection'' studies of brain activity. Table \ref{my-label} summarizes our work against some other real-fake detection studies.

}

\setlength{\parskip}{0cm}
    \setlength{\parindent}{1em}
\section{Study Design \& Data Collection}
\label{Design of Experiments}
 

\subsection{Ethical and Privacy Considerations}
{Our study was approved by the Institutional Review Board (IRB) at our university. Participation in the study was completely voluntary and followed informed consent. The participants were given a chance to leave the study at any point if they did not feel comfortable. We utilized the best practices to make sure the participants' private data (such as the survey results, and the neural data obtained during the experiment) was protected and duly anonymized.}

\subsection{Design of the News Credibility Detection Task}
{We designed an experimental task in which the participants were asked to read the different (real and fake) news articles while wearing the MindWave Mobile 1 headset, which recorded their brain activity during the task. Each participant was asked to read 40 articles (20 fake and 20 real). The articles were drawn from the dataset which is used in a study conducted by Zannettou et al. \cite{ZannettouCCKLSS17}. The dataset consists of 99 news sites gathered from millions of comments, threads, and posts on Twitter, Reddit, and 4chan. These websites\footnote{The list of the 99 sites is available at \url{https://drive.google.com/file/d/0ByP5a__khV0dM1ZSY3YxQWF2N2c/view?usp=sharing\&resourcekey=0-nyqLTloH-4PFfOscFB0G1Q}} including 45 mainstreams (characterized as real news) from the Alexa top 100 news sites and 54 alternative websites (characterized as fake news) from Wikipedia\footnote{\url{https://en.wikipedia.org/wiki/List_of_fake_news_websites}} and FakeNewsWatch \cite{fakenewswatch} were gathered from posts, threads, and comments on three platforms. We used this dataset as follows. Due to a lapse of time, we first tested if the alternative domains were working and realized that 24 domains had already expired. From the remaining 75 domains, we randomly, but considering the length of the news (readable in 30 seconds), chose 200 news including 100 fake news, and 100 real news to create our small dataset. Then, we again randomly chose 20 fake news articles and 20 real news articles from our small dataset. We chose the same number of articles from different categories such as politics, daily, and other (sports, science, business, entertainment, etc.) news.

Since our goal was to assess the credibility of the news
articles themselves, we only showed the textual content of these news
 articles to the participants, rather than the side information such as URLs and
 advertisements. We believe this aspect of our study design is crucial and in fact, a strength of our work because, as demonstrated in 
most phishing attack studies \cite{Dhamija:2006, neupaneEEG}, people usually do not understand or heed the URLs. Further, fake news is often hosted on legitimate websites for which the users already trust the corresponding URLs, and fake news frequently gets propagated via users' social media accounts with textual postings of the news content.

In the study, the articles were pre-downloaded for offline usage and hosted on a local webserver to be displayed using the Firefox browser. We modified the shown page with a white background with black text and in the same style for all articles. We also removed the images from the original news articles (if any). The URLs were hidden during the experiment. After reading each article, the participants were asked to identify whether it is real or fake.

The exact flow of the experiment was as follows. First, the instruction page, which includes the directions of the experiment, was shown for 20 seconds and the experiment start page informed the participants about the beginning of the experiment. Then, the resting page, which is for participant relaxation and eliciting baseline brain activity, was shown for 2 seconds. Right after this, the trial (fake or real) news article was displayed for 30 seconds. 30 seconds represent a sufficient amount of time for the participants to read/analyze the news article as they were of average to short length. The order of presentation of the news articles was randomized per participant. After presenting each news trial, we asked the participants, “Do you think the shown news is real?” with “Yes” and “No” answer choices. Finally, the goodbye page was displayed to end the experiment for 5 seconds. Figure \ref{FlowChart} visualizes the experimental flow.
}
\begin{figure}[h!]	
	\centering
	\vspace{-3mm}
	\includegraphics[scale=.46]{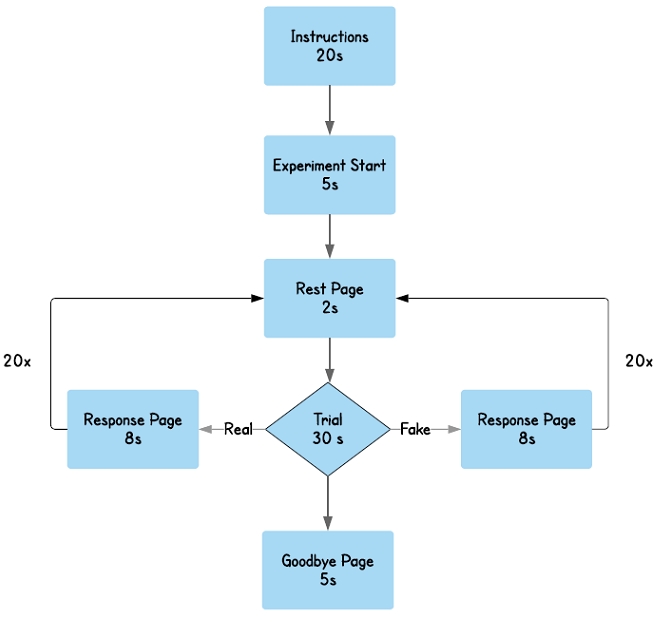}	
	\caption{The flow diagram represents design of the experiment.}
	\vspace{-1mm} 
	\label{FlowChart}
\end{figure}
\vspace{-2mm}
\subsection{Experimental Set-Up}
{In our experiment, participants were asked to wear the EEG headset on their head. While they performed the experimental task, we recorded their brain activity in terms of EEG frequency bands, metrics and raw signals (defined in section \ref{sec:back}). For the experimental setup, we developed Phyton scripts to read EEG data using socket. For this experiment we used PC monitor with screen size of 1920*1080. Along with the EEG data, we recorded the timestamps of while news articles were showed and when our participants were resting.
}
\subsection{Study Protocol}
\label{(study-protocol)}
{\noindent \textbf{Participant Recruitment and Preparation Phase:} 
	{
	Twenty-three healthy people (mainly students and workers at our University) participated in our study. After we eliminated the polluted data samples containing noisy brain signals (due to participants moving excessively), the final sample of participants came out to be nineteen. Each participant completed the study in about 30 minutes. After giving consent information, the participants were asked to provide their demographic information (such as gender, education level, and current job).
	Many of our participants were in the range of 18-30 years (58\%) and male (58\%). Most of them had at least a bachelor’s degree (84\%) and 32\% had a job currently. Table \ref{particpantdemographics} provides a summary of the demographic information of the participants. Previous power analysis studies have found 15 to be an
optimal number of participants for such studies. For example,
statistical power analysis of ER-design fMRI studies has
demonstrated that 80\% of clusters of activation proved reproducible
with a sample size of 15 subjects \cite{murphy2004}. Our participant demographics is also well-aligned with prior
neuroimaging security studies \cite{andersen-fmri,neupaneEEG,neupanefNIRS,neupane2014neural}.}

\begin{table}[h!]
	\scriptsize
	\centering
	\caption{Summary of Participant Demographics}	
	\begin{tabular}{|l|l|r|}
		\hline
		\textbf{Demographics (N=19)}       &                                                                                                        & \textbf{\%}                                                       \\ \hline \hline
		 
		\textbf{Gender}                    & \begin{tabular}[c]{@{}l@{}}Male\\ Female\end{tabular}                                                  & \begin{tabular}[c]{@{}r@{}}57.9\\ 42.1\end{tabular}               \\ \hline
		 
		\textbf{Age}                       & \begin{tabular}[c]{@{}l@{}}18-24\\ 25-30\\ 31-40\\ \textgreater{}40\end{tabular}                       & \begin{tabular}[c]{@{}r@{}}26.3\\ 31.6\\ 21.1\\ 21.0\end{tabular} \\ \hline
		
		\textbf{Education}                 & \begin{tabular}[c]{@{}l@{}}Bachelor's degree\\ Master's degree\\ Doctoral degree\\ Others\end{tabular} & \begin{tabular}[c]{@{}r@{}}52.6\\ 26.3\\ 5.3\\ 15.8\end{tabular}  \\ \hline
		
		\textbf{Employment status}        & \begin{tabular}[c]{@{}l@{}}Employed\\ Non-Employed\end{tabular}                                        & \begin{tabular}[c]{@{}r@{}}31.6\\ 68.4\end{tabular}               \\ \hline
		
		\textbf{Frequency of Reading news} & \begin{tabular}[c]{@{}l@{}}Daily\\ Several times a week\\ Once a week\\ Others\end{tabular}            & \begin{tabular}[c]{@{}r@{}}21.1\\ 47.4\\ 21.1\\ 10.4\end{tabular} \\ \hline
		\textbf{Credibility of the News}   & \begin{tabular}[c]{@{}l@{}}Care\\ Do not care\end{tabular}                                             & \begin{tabular}[c]{@{}r@{}}73.7\\ 26.3\end{tabular}               \\ \hline
	\end{tabular}
	
	\label{particpantdemographics}
	\vspace{-1mm}  
\end{table}

	\noindent \textbf{Task Execution Phase:} 
	{To recall, NeuroSky’s Mobile 1 Headset was used to collect the EEG data for the experiment. We placed the headset on the head of each user, then started to run the experiment and record the brainwaves simultaneously.
The ThinkGear software (Section \ref{eegoverview}) was used as an interface. This software included real-time artifact removal due to muscle movement and environment/electrical interference such as spikes, saturation, and movement of the head.}
	
	\noindent \textbf{Post-Experiment Phase:} 
	{Once the experiment was completed, the post-experiment survey page,
	designed to learn participants' interests in news articles and their strategies to assess their credibility, loaded into the browser. 68\% of the participants said they read news from the news websites, 
and 89\% participants read at least a news during the week.
The most common answer to the question of caring about news credibility was “Yes” (73\%). Upon asking the participants what their strategy was to identify whether the news is real or fake, one of the participants said that when the news is about an opinion or a reporter's thinking, he does not believe in the news story. Some of them claimed that when the news sounds unreal, they prefer to verify from several sources. 
}

\section{Analysis Methodology}
\label{analysis}
{
}

\noindent \textbf{Processing of EEG Bands and Metrics:}
\label{process neural data}
{	
We collected continuous data of raw EEG signals, the frequency band information of the signals (Alpha, Beta, Gamma, etc.), and Neurosky's post-processed, aggregated EEG metrics (Attention and Meditation). Three different trials were considered in our recorded EEG data: (1) the fake news, (2) the real news, and (3) the resting state that was for the participants' relaxation between the other trials. 
We set up 2 seconds for the resting state, 30 seconds for each real/fake news trial, and 5 seconds for response time and chopped the neural data per trial and the resting state time using the onset and offset time for each. So, we have 40 data samples corresponding to the trials and the resting state for each session.
For the fake vs. real news trial analysis, we calculated the average of the frequency bands and metrics for each trial which has multiple rows of data. The same process was followed for news trials vs. the resting state analysis.}

\noindent \textbf{Processing of Raw Data:}
\label{process raw data}
{
\textit{ThinkGear Socket Protocol} \cite{thinkgearsocket} captures the raw EEG signal in the time domain. To analyze the differences between real and fake news trials, and between trials and the resting state, we converted the signals from the time domain to the frequency domain. To this end, we used Signal Analyzer Application (SAA) \cite{signalanalyzerapp} in MATLAB \cite{matlab}, a tool to visualize and analyze the data by comparing the signals in both domains. For each case, a high pass filter (HPF) and a low pass filter (LPF) were applied separately. The passband frequency was defined as 0.5 for both filters. The main difference between a high-pass filter and a low-pass filter is that a low-pass filter allows the signal to pass through lower than a cutoff frequency, while a high-pass filter allows the signal to pass through higher than a cutoff frequency. In other words, a low-pass filter blocks high frequencies, and a high-pass filter blocks low frequencies.
}

\noindent \textbf{Statistical Testing Methodology:} 
{
For the statistical analysis, we did two types of comparisons using IBM's SPSS Statistical Software \cite{IBMSPSS}. One for fake vs. real news trial and another for news trial (fake and real) vs. resting state. 
All statistical results in this paper are reported at a significance level $(\alpha)$ of 0.05. The Friedman test was used to test for the existence of differences within the above groups, and the Wilcoxon Signed-Rank Test (WSRT) was used for measuring the pairwise differences between different EEG bands and metrics underlying our analysis. The effect size of WSRT was calculated using the formula $r=Z/\sqrt N$, where $Z$ is the value of the z-statistic and $N$ is the number of observations on which $Z$ is based. We considered Cohen criteria \cite{cohen1977statistical} which reports effect size $>$ .1 as small, $>$ .3 as medium and $>$ .5 as large. The statistically significant pairwise comparisons are reported with Bonferroni corrections. 
}
\section{Task Performance Results}
\label{behavioral-results}
{

In the first phase of our analysis, we calculated the response time and the fraction of correctly identified articles out of the total number of responses given by the participants (termed ''accuracy'') for the real and fake news trials. The goal of this analysis was to determine how accurately the participants were able to identify two types of trials. Table \ref{tab:tasktab} summarizes the result from our task performance analysis.
\begin{table}[h]
\center
\scriptsize
\caption{Task Performance in Identifying Fake and Real News Articles}
\begin{tabular}{|l|l|l|l|}
\hline
{\textbf{Trial}} & {\textbf{Accuracy (\%)}} & {\textbf{Wrong Response (\%)}} & {\textbf{Response Time (ms) $\mu$($\sigma$)}} \\ \hline \hline
{\textbf{Real}} & 54.41 & 45.29 & 3069.4 (9.33) \\ \hline \hline
{\textbf{Fake}} & 51.76 & 48.24 & 2966.7 (9.34) \\ \hline
\end{tabular}
\label{tab:tasktab}
\vspace{-1mm}

\end{table}

We calculated the accuracy of the detection task
for two types of responses: one for identifying correct responses (real as real and fake as fake), and another for wrong responses (real as fake, fake as real). So, in total there are four types of comparisons for decision making: real as real, real as fake, fake as fake, and fake as real.

From Table \ref{tab:tasktab}, we can see that the overall accuracy of our participants in the experiment is 53.09\%. Here,  we can also see the accuracy for real news articles is 54.41\% and for fake news articles is 51.76\%. This means that the accuracy of the identification of real news articles as real news is almost similar to the accuracy of the identification of fake news articles as fake news. Also, if we look at the wrong responses, the percentage of identification of real news articles as fake news is 45.59\% and fake news articles as real news is 48.24\%, and the total percentage of wrong answers is 46.91\%. Indeed, upon using WSRT, we did not notice any statistically significant differences in the mean of the overall accuracy of real and fake articles. The average response times for real and fake trials are also closely similar and not statistically significant. 

These results suggest that participants did not at all do well in telling the differences between real and fake news articles and their accuracy in identifying the two types of articles is very close to a random guessing accuracy (50\%).  
}

\setlength{\parskip}{0cm}
    \setlength{\parindent}{1em}
\section{Neural Results}

\label{Results and Analysis}
In this section, we investigate the neural activation when users were reading and deciding between the real and fake articles. In particular, we compare the neural activity between news (fake and real) trials and the baseline condition (resting state), and between the real news trials and fake news trials.

\subsection{Trial vs.~ Resting State}
\label{trial-rest}
{To evaluate the
brain activity levels when participants were reading different articles and
assessing their credibility, we first contrasted the brain activation during
real and fake news trials with the resting state as a ground truth.

\begin{table}[h!]
	\centering
	\scriptsize
	\caption{Statistically Significant Results across all news articles.}
	\vspace{-3mm}
	\label{tab:stattab}
	\smallskip
\begin{tabular}{|l|l|r|r|}
\hline
\textbf{Comparison}                              & \textbf{Metrics} & \textbf{p-value} & \textbf{Effect Size} \\ \hline \hline
\multirow{3}{*}{\textbf{Real \textgreater Rest}} & Delta            & 0.001            & 0.53 (large)        \\ \cline{2-4}
                                                 & Attention        & 0.001            & 0.62 (large)         \\ \cline{2-4}
                                                 & Meditation       & 0.001            & 0.52 (large)         \\ \hline \hline
\textbf{Fake \textgreater Rest}                  & Attention        & 0.001            & 0.56 (large)       \\ \hline
\end{tabular}
\vspace{-2mm}
\end{table} 

As described in Section \ref{process neural data}, we extracted and calculated the frequency band information and the metrics collected from the EEG data. From Table \ref{tab:meanstddtab}, we can see the differences between the mean (and standard deviations) of news trials and the resting state. We observe that the means of Delta, Theta, and Low Alpha of news trials are lower than the corresponding means of the resting state. The means of High Beta, Low Gamma, High Gamma, Attention, and Meditation are lower in the resting state than in the news trials, but the means of High Alpha and Low Beta in the resting state are close enough within the range of real and fake news.

To confirm whether these differences were statistically significant, we ran the
Friedman test in two groups; one for EEG bands (Delta, Theta, Low Alpha,
High Alpha, Low Beta, High Beta, Low Gamma, High Gamma) and another for EEG metrics
(Attention and Meditation), which indeed revealed a statistically significant
difference (${\chi}^2$ = 331.269, p = 0.000) corresponding to news trials and
resting state. Subsequently, we ran WSRT with
Bonferroni correction to evaluate the pairwise differences in the mean of each of
the EEG frequency bands and the metrics between news trials vs. the resting
state. We noticed that only Attention is statistically significant between the
fake news trials and resting state, and Attention, Meditation and Delta are
statistically significant between the real news trials and resting state.
The statistically significant results are depicted in Table \ref{tab:stattab}. From Table
\ref{tab:meanstddtab}, we can see that the raw mean values of Attention (51.96) in
fake news trials are greater than the mean of the resting state (46.08),
which means participants were more attentive to reading fake news articles than
they were in the resting state. In comparing real news trials vs. resting
state, we can see that the mean value of Attention (52.44) in the real news trial
is higher than the resting state (46.08), mean of Meditation (56.16) in the real trial is higher than the mean of the resting state (53.12). Finally, the mean of the Delta
band in the resting state (502423.87) is higher than the mean of Delta in the real
trial (400910.12).

\begin{table*}[ht!]
	\small
	\caption{Overall Neural Activation Results for Real News Trials, Fake News Trials and Resting State (Baseline)} \label{tab:meanstddtab}
	\scriptsize
\begin{tabular}{|l|l|r|r|r|r|r|r|r|r||r|r|}
		\hline
		\multicolumn{2}{|l}{\textbf{Overall}}  & \multicolumn{8}{|c||}{\textbf{EEG Bands}} & \multicolumn{2}{|c|}{\textbf{EEG Metrics}} \\ \hline
		\hline
		\textbf{Type}&\textbf{Value}  & \textbf{Delta} & \textbf{Theta} & \textbf{LowAlpha}   & \textbf{HighAlpha} & \textbf{LowBeta} & \textbf{HighBeta} & \textbf{LowGamma} & \textbf{HighGamma} & \textbf{Attention} & \textbf{Meditation} \\ \hline \hline
		
		\textbf{Real}    & $\mu$     & 400910.12&	88544.69&	29227.06&	26886.87&	24506.83&	18795.82&	9675.33&	5889.36&	52.44&	56.16 \\ \hline
		& $\sigma$ &147104.46&	24040.17&	7511.16&	5739.33&	6190.91&	6069.21&	3832.33&	1972.76&	4.41&	5.52\\ \hline \hline
		
		\textbf{Fake}    & $\mu$     & 414988.46 &95187.77&29177.07&27430.93&24723.57	&18949.93&9382.27&5883.22&51.96	& 55.78  \\ \hline
		& $\sigma$ &150213.95&	26768.06&	8255.18&	5754.46&	6405.26&	5207.92&	3396.58&	2204.92&	4.23&	5.11 \\ \hline \hline
		
		\textbf{Rest}    & $\mu$     & 502423.87&	110839.71&	31494.11&	27029.51&	24643.74&	17197.73&	9094.13&5653.64&46.08&53.12 \\ \hline 
		& $\sigma$ &150741.06&	36842.97&	11718.16&	9015.74&	9719.48&	7541.49&	3821.34&	2429.54&	4.84&	7.06\\ \hline
	\end{tabular}	
\vspace{-2mm}
\end{table*}

\begin{table*}[ht!]
	\centering	
	\scriptsize
	\caption{Categorical Neural Activation Results for Real News Trials, Fake News Trials and Resting State (Baseline)}
	\begin{subtable}
		\centering
		\begin{tabular}{|l|l|r|r|r|r|r|r|r|r||r|r|}
		\hline
		\multicolumn{2}{|l}{\textbf{Politics}}  & \multicolumn{8}{|c||}{\textbf{EEG Bands}} & \multicolumn{2}{|c|}{\textbf{EEG Metrics}} \\ \hline
			\hline
			\textbf{Type} &\textbf{Value} & \textbf{Delta} & \textbf{Theta} & \textbf{LowAlpha}   & \textbf{HighAlpha} & \textbf{LowBeta} & \textbf{HighBeta} & \textbf{LowGamma} & \textbf{HighGamma} & \textbf{Attention} & \textbf{Meditation} \\ \hline \hline
			
			\textbf{Real}    & $\mu$       & 413320.05&	96793.34&	29878.77&	26694.87&	25268.51&	19153.27&	9901.83&	5979.33&	52.20&	54.24 \\ \hline 
			& $\sigma$ &153757.50&	34383.43&	8605.00&	6055.53&	7050.92&	7295.16&	3803.14&	2140.09&	5.43&	6.77\\ \hline \hline
			
			\textbf{Fake}    & $\mu$     & 424955.60&	95244.30&	28988.54&	27138.52&	24732.95&	19065.21&	9528.1400&	5829.22&	51.98&	55.53  \\ \hline 
			& $\sigma$ &140836.10&	23006.76&	6495.04&	5142.80&	5670.08&	4809.90&	3129.81&	1895.49&	4.10&	4.20 \\ \hline\hline
			
			\textbf{Rest}    & $\mu$       & 491203.76&	103688.32&	30913.01&	24676.13&	22601.05&	15977.85&	8534.58&	5346.13&	46.31&	53.06 \\ \hline 
			& $\sigma$ &184290.10&	26949.64&	11145.64&	7143.12&	6854.26&	4167.79&	2996.65&	1721.91&	4.76&	7.64\\ \hline
		\end{tabular}
	\end{subtable}
\newline
	\vspace*{-0.3 mm}
	\newline
	\begin{subtable}
		\centering
		\begin{tabular}{|l|l|r|r|r|r|r|r|r|r||r|r|}
		\hline
		\multicolumn{2}{|l}{\textbf{Daily}}  & \multicolumn{8}{|c||}{\textbf{EEG Bands}} & \multicolumn{2}{|c|}{\textbf{EEG Metrics}} \\ \hline
			\hline
			\textbf{Type} &\textbf{Value} & \textbf{Delta} & \textbf{Theta} & \textbf{LowAlpha}   & \textbf{HighAlpha} & \textbf{LowBeta} & \textbf{HighBeta} & \textbf{LowGamma} & \textbf{HighGamma} & \textbf{Attention} & \textbf{Meditation} \\ \hline \hline
			
			\textbf{Real}    & $\mu$      & 406622.23&	87742.71&	28911.04&	27968.43&	24659.73&	18860.05&	9924.38&	5888.73&	51.79&	56.28 \\ \hline
			& $\sigma$ &168881.12&	24040.13&	7871.16&	7434.02&	8012.35&	7785.45&	5320.73&	2385.62&	4.86&	6.73\\ \hline\hline
			
			\textbf{Fake}    & $\mu$       & 415203.58&	103724.81&	31690.43&	27770.25&	24805.17&	19355.21&	9983.93&	6466.82&	47.86&	54.68  \\ \hline
			& $\sigma$ &177150.56&	40763.64&	16106.40&	8980.59&	9594.85&	10168.33&	5387.70&	3238.25&	8.07&	8.04 \\ \hline\hline
			
			\textbf{Rest}    & $\mu$      & 490922.25&	105614.17&	28735.76&	24745.74&	22367.27&	15297.24&	8118.90&	5262.25&	45.74&	51.17 \\ \hline
			& $\sigma$ &220816.90&	44285.52&	12952.37&	7534.54&	8780.23&	5165.82&	3537.03&	2485.51&	6.06&	7.46\\ \hline
		\end{tabular}
	\end{subtable}
\newline
	\vspace*{-0.3 mm}
	\newline
	\begin{subtable}
		\centering
		\begin{tabular}{|l|l|r|r|r|r|r|r|r|r||r|r|}
		\hline
		\multicolumn{2}{|l}{\textbf{Other}}  & \multicolumn{8}{|c||}{\textbf{EEG Bands}} & \multicolumn{2}{|c|}{\textbf{EEG Metrics}} \\ \hline
			\hline
			\textbf{Type} &\textbf{Value}  & \textbf{Delta} & \textbf{Theta} & \textbf{LowAlpha}   & \textbf{HighAlpha} & \textbf{LowBeta} & \textbf{HighBeta} & \textbf{LowGamma} & \textbf{HighGamma} & \textbf{Attention} & \textbf{Meditation} \\ \hline \hline 	
			
			\textbf{Real}    & $\mu$      & 388485.37&	83570&	29070.39&	26495.37&	24318.91&	18781.26&	9333.63&	5835.77&	53.12&	57.61 \\ \hline
			& $\sigma$ &143754.05&	21048.06&	8485.08&	6372.84&	7233.89&	6719.034&	4064.91&	2492.71&	6.43&	6.08\\ \hline\hline
			
			\textbf{Fake}    & $\mu$      & 408582.25&	93892.00&	29007.76&	27672.92&	24758.34&	18913&	9231.12&	5877.95&	52.76&	56.18  \\ \hline
			& $\sigma$ &167840.62&	31452.57&	9296.78&	6819.27&	7206.02&	5680.82&	3876.19&	2814.76&	5.29&	6.45 \\ \hline\hline
			
			\textbf{Rest}    & $\mu$     & 515457.83&	118797.9&	33269.67&	30050.28&	27418.84&	19130.14&	10013.07&	6081.46&	46.13&	54.28 \\ \hline
			& $\sigma$ &148661.33&	57555.89&	15874.06&	14312.91&	15000.57&	9926.36&	5587.18&	3681.14&	4.99&	8.06\\ \hline
		\end{tabular}
	\end{subtable}
\vspace{-0.8mm}
\newline
	\label{tab:catmeantab}
\end{table*}

From Table \ref{tab:frequencybands},
we learned that Attention represents the concentration of
awareness on some phenomenon and Meditation represents
mental calmness \cite{meditation}. In fake-rest pair,
Attention is high in the fake state, thus we may conclude that participants may
have been more concentrated when reading fake news articles than in the resting state.
Also, in the real-rest pair, Attention and Meditation are high in the real trial than
resting state. This suggests that participants were reading/assessing the
real news articles with full awareness but they were relaxed while reading such
articles. This may mean participants found the real news articles
related to real-life events, thus their brain signals implicated calmness. In
the real-rest pair, Delta is also higher in the resting state. It seems to suggest that
participants were not conscious while resting. As Delta represents deep
dreamless sleep and unconsiousness\cite{hirnwellen},\cite{brainmapping}, this
result could mean that the participants were as relaxed as dreamless sleep (in
slow-sleep dreaming may also occur, but dreamless sleep is like deep meditation).
This analysis, therefore, suggests that participants have been more engaged and
less distracted when reading and analyzing the news articles than during the
resting state. This also serves to confirm that the experiment worked as desired as one would
expect the participants to be significantly more active during the news trial
presentation than during resting.}

\subsection{Real News Trial vs.~ Fake News Trial}
\label{realvsfake}
{
Having established that both real and fake news trials invoked significantly higher brain activity compared to the resting state, we now set out to analyze potential differences in the brain activation between the real news trials and the fake news trials. These comparisons of brain activities could help delineate the neural processing involved in reading and identifying real and fake articles. To do this analysis, we compared the mean and standard deviation corresponding to different EEG bands and metrics for fake news trials and real news trials. Our results are presented in Table \ref{tab:meanstddtab}.

Just looking at the raw mean values from this table, the mean of Delta, HighAlpha, HighBeta, and HighGamma are slightly higher in fake articles, but these do not seem to be significantly higher. Also, the mean value of LowBeta for fake articles and real articles is approximately the same. On the other hand, the mean of the Theta band seems significantly higher in fake articles. 

The Friedman test revealed a statistically significant difference in the means of the group of different EEG bands (Delta, Theta, lowAlpha, highAlpha, lowBeta, highBeta, lowGamma, highGamma) and in the mean group of EEG metrics (Attention and Meditation) (${\chi}^2$ = 664.269, p = 0.000) corresponding to real vs. fake news trials. Upon applying WSRT (\textit{without} Bonferroni correction) to find the pairwise differences in the means of the different EEG bands and metrics corresponding to real and fake news trials, the p-value for Theta turned out to be .049, which represents a statistically significant difference (rest of the bands and metrics did not show any statistical difference). However, crucially, \textit{after applying} the Bonferroni correction, this only significant difference between the Theta band of real and fake news trials disappears. Therefore, our overall analysis shows that there is \textit{no statistically significant difference} in any of the EEG bands and EEG metrics corresponding to real and fake news trials.

The analysis suggests that our participants were likely equally engaged and relaxed in reading and identifying real and fake news articles, i.e., their brains did not seem to react differently when processing real and fake news articles.}

\subsection{Categorical Analysis}
\label{Categorical Analysis}

Now we analyze the neural activation based on different category news articles. First, we compared the neural activation between real or fake trials and the resting state, then we compare the neural activation between real trials and fake trials, and finally, we compare the neural activation for news trials between categories.

\noindent \textbf{Political News:} In political news, raw mean values, depicted in Table \ref{tab:catmeantab}, for the metrics and bands are approximately the same in real and fake news trials. Thus, we do not see any significant differences in the mean values for real trials and fake trials. However, the resting state shows a different result in Delta, Theta, HighAlpha, LowBeta, LowGamma, HighGamma, and Attention metrics. Meditation has a slightly different value but is not significantly different. To confirm the difference, we ran the Friedman test and it revealed statistically significant differences in both the mean group of EEG bands and EEG metrics corresponding to this category. Upon performing pairwise comparisons using WSRT (with Bonferroni correction), as depicted in Table \ref{tab:category2}, we can find that Attention and HighBeta are statistically significant in the fake-rest pair and Attention is statistically significant in the real-rest pair. However, we did not find any statistically significant differences between real news trials and fake trials in this category.  From Table \ref{tab:catmeantab}, we noticed that the raw mean values of Attention and HighBeta are lower in the resting state than in the fake news trial, and the means of Attention are slightly lower in the resting state than in real news trial.

This analysis shows that the participants were reading fake political news articles with full awareness and the result represents nervous excitement as Attention and HighBeta are higher in the fake news trial than resting state. Also, the higher Attention level in the real news trial represents that the participants were reading real political articles with full awareness. However, nervous excitement or the state of anxiety exists while reading fake political articles. While news trials differ significantly from the resting state, no differences emerged between real news trials vs. fake news trials.

\noindent \textbf{Daily News:} For daily news articles, we noticed from Table \ref{tab:catmeantab} that the raw mean values of Theta, LowAlpha, HighGamma, Attention, and Meditation are slightly different between fake news trial and real news trial. Resting-state is also showing different results for Delta, Theta, LowBeta, HighBeta, LowGamma, HighGamma, Attention, and Meditation. Following the same pattern of statistical analysis as in the Politics news category, however, we did not see any statistical difference in the real-fake pair and fake-rest pair, but in the real-rest pair Attention and Meditation are statistically significant. From the mean values in Table \ref{tab:catmeantab}, we do see that Attention and Meditation are higher in real trials than resting state. 

This analysis, therefore, reveals that there are no differences in real and fake metrics in daily news articles, but Attention and Meditation are significantly different in resting state in the real-rest pair. As Attention represents awareness and meditation represents mental calmness, we can infer that the participants were reading the real daily news articles with full awareness but their brain activation also represents relaxation.

\noindent \textbf{Other News:} We followed the same pattern of analysis for the ``Other'' category of news and noticed that, via eyeballing the mean values, the real and fake news trials have an approximately similar output for all bands and metrics, whereas the resting state has different result compared to the fake and real news trials. Indeed, the statistical analysis supported this observation. We found that Attention is statistically significant in fake-rest pairs and Delta is statistically significant in real-rest pairs. According to the mean values from Table \ref{tab:catmeantab}, we do see that mean of Attention in fake trials (52.76) is higher than the mean of Attention (46.13) in resting states. Thus, participants were more attentive to reading other news articles than the resting state. The mean of Delta is higher in the resting state than the mean of Delta in real trials. This result shows that the participants were more attentive to reading other fake news articles than the resting state. Also, the higher value of Delta in the resting state of the real-rest pair represents the relaxation, unconsciousness, and a state of dreamless sleep (to recall, in slow-sleep dreaming may also occur, but dreamless sleep is like deep meditation) \cite{hirnwellen},\cite{brainmapping}).

We can conclude that the participants were more relaxed in their resting state than in trials for both pairwise comparisons of the other news category (real-rest and fake-rest). Also, the brain activation did not show any differences between real and fake news. Thus the participants may not have been able to distinguish the real and fake other news articles but their brains reacted differently when they were assessing news articles and taking a rest in the resting state.

\begin{table}[h!]
	\scriptsize	
	\centering
	\caption{Statistically significant results per category of news articles (No statistically significant differences were observed)}
 \label{tab:category2}
 \vspace{-1mm}
\begin{tabular}{|l|l|l|r|r|}
\hline
\textbf{Categories}                & \textbf{Comparison}                              & \textbf{Metrics} & \textbf{p-value} & \textbf{Effect Size} \\ \hline \hline
\multirow{3}{*}{\textbf{Politics}} & \multirow{2}{*}{\textbf{Fake \textgreater Rest}} & High Beta              & 0.003            & 0.49 (medium)         \\ \cline{3-5} 
                                   &                                                  & Attention              & 0.001            & 0.57 (large)         \\ \cline{2-5} 
                                   & \textbf{Real \textgreater Rest}                  & Attention              & 0.002            & 0.51 (large)         \\ \hline \hline
\multirow{2}{*}{\textbf{Daily}}    & \multirow{2}{*}{\textbf{Real \textgreater Rest}} & Attention              & 0.001            & 0.54 (large)         \\ \cline{3-5} 
                                   &                                                  & Meditation             & 0.001            & 0.55 (large)         \\ \hline \hline
\multirow{2}{*}{\textbf{Other}}    & \textbf{Fake \textgreater Rest}                  & Attention              & 0.001            & 0.60 (large)         \\ \cline{2-5} 
                                   & \textbf{Real \textless Rest}                     & Delta                  & 0.003            & 0.48 (medium)        \\ \hline
\end{tabular}
\vspace{-2mm}
\end{table}

\noindent \textbf{Between-Category Comparison:} We performed a pairwise comparison of each trial/resting state between all categories (Example pairs: real-daily vs. real-politics, fake-daily vs. fake-politics, rest-daily vs. rest-politics, etc.). We followed the same pattern of analysis mentioned above to compare these pairs between every two categories. We found that the Attention metric is statistically significantly different in other-daily pairs corresponding to fake news trials (p-value of 0.018). From Table \ref{tab:catmeantab}, we can indeed see the mean of Attention (52.76) is higher in Other Type than the mean of Attention (47.86) in daily news for fake trials. As Attention represents awareness, it means the participants were more attentive to reading other fake news articles than the daily fake news articles. This may suggest that the participants were not reading daily news fake articles with full awareness as those articles may look less interesting or the participants may have identified those articles as fake, so they were not reading them seriously.
\subsection{Raw Data Analysis}
\label{Raw Data Analysis}
Our analysis of the EEG bands and EEG metrics showed that there was no statistically significant difference between real and fake news trials. Since the bands and metrics are obtained from the post-processing over raw EEG signals, this information may have missed some characteristics of the brain activity that might be different in the real and fake news trials. To explore any such potential differences, we performed additional analysis directly over the raw EEG signals corresponding to real vs. fake news trials.

\begin{figure}[h!]	
	\centering		
	\begin{subfigure}{}
	\centering
		\includegraphics[width=0.45\columnwidth]{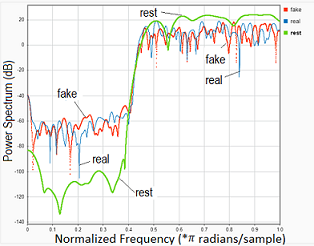}		
		\label{fig:highpass}			
	\end{subfigure}%
	\begin{subfigure}{}
	\centering
		\includegraphics[width=0.45\columnwidth]{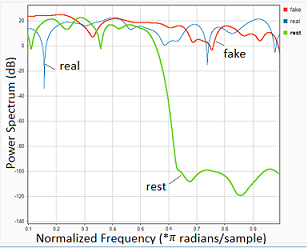}		
		\label{fig:lowpass}		
	\end{subfigure}
	\vspace{-2mm}
	\caption{Average of all participants' EEG raw signals frequency\\ High-pass filtered (left), Low-pass filtered (right)}
	\vspace{-1mm}
	\label{rawfig}	
\end{figure}

We followed the methodology presented in Section \ref{process raw data} to analyze the raw EEG signals. When we first look at the average of the frequency for both high-pass filtered and low-pass filtered results in Figure \ref{rawfig}, we found significant differences between the news trial (fake or real) and the resting state. Further, the power spectrum values were -80 dB and -120 dB for the resting state, while the real/fake news trials were between -40 dB and -100 dB in the range of 0 to 0.45 frequency after applying the high-pass filter (Figure \ref{rawfig} (top)). Similarly, in the frequency range of 0.65 to 1.00 on low-pass filtered signals Figure \ref{rawfig} (bottom), the minimum resting-state value was -100 dB, and the maximum resting-state value was  -120 dB, however, the fake and the real news trials were in the range of -30 dB and 20 dB.

These results suggest that especially after applying a high-pass filter to the signals, there were no observable significant differences between the fake and the real news trials. Likewise, low-pass filtered signals are sloping similarly in terms of fake and real news trials. On the other hand, the difference between the real/fake news trials and the resting state is significant for both high-passed signals and low-passed signals. This means that the participants were actively engaged in reading and deciding between the real and fake news articles, but their brains were not processing real and fake news articles differently, even when looking at the raw neural signals.

\section{Brain Signal Analytics: \\Fake vs. Real News Classification}
\label{auto}

Statistically, we did not observe differences in users' neural activities in the fake vs. real news identification task.
In this section, we classify the real vs. fake news neural activity data using machine learning algorithms
for further validation of the statistical result. 
\subsection{Features and Classification Algorithms}
\label{feature-extraction}

We normalized the data obtained from all the 19 participants using the average
of each of the EEG bands and EEG metrics for each participant. Then, we
calculated various statistical features, i.e., the maximum, minimum, mean, standard deviation, variance, skewness,
and kurtosis values, for each participant from this normalized data. With these
feature values, we created a new data set to be used as an input to machine-learning
algorithms. In this dataset, EEG bands (Delta, Theta, Low Alpha, High Alpha, Low
Beta, Low Beta, Low Gamma, High Gamma) and EEG metrics (Attention and Meditation)
values were defined separately for fake and real news trials.

We utilized the off-the-shelf algorithms provided by Weka
\cite{Witten:2011:DMP:1972514} to build the classification models and test them
using 10-fold cross-validation. The algorithms we tested are 
follows: J48, Random Forest (RF), and Random Tree (RT) under
the \textit{Trees} category of models, Neural Networks, Multilayer Perceptron (MP), Support
Vector Machines (SMO) and Logistics (L) under the \textit{Functions} category and
Naive Bayes (NB) under \textit{Bayesian Networks} the category.

\subsection{Classification Results} 
\label{defensive-mechanism}

For the classification task, the positive class was defined as real news and the negative class
as fake news. In terms of classification performance metrics, we computed precision (\textit{Prec}), Recall
(\textit{Rec}), F-measure (\textit{FM}), \textit{True positive} (TP) and
\textit{False positive} (FP) values and report on the average of these values.
\textit{Prec} value shows the accuracy of the system in rejecting negative
classes and evaluates the security of the tested approach (i.e., rejecting fake news as invalid news). On the other hand,
the \textit{Rec} value evaluates the usability of the approach in accepting
positive classes (i.e., accepting real news as valid news). \textit{FM} shows the balance between precision and recall.
TP represents the correctly defined total rate of positive classes (real news), while FP
represents the incorrectly defined total rate of positive classes.

\begin{table}[ht!]
\centering
\scriptsize
\caption{Fake-Real News Classification}
\begin{tabular}{|p{3cm}|l|l|l|l|l|} 
\hline
\textbf{}        & \textbf{TP Rate} & \textbf{FP Rate} & \textbf{Prec} & \textbf{Rec} & \textbf{FM}  \\ 
\hline
\textbf{All Features}   & \multicolumn{5}{l|} {} \\
\hline
RandomTree                   & 0.47             & 0.53             & 0.47          & 0.47         & 0.47         \\ 
\hline
Logistic                     & 0.31             & 0.69             & 0.31          & 0.31         & 0.31         \\ 
\hline
J48                          & 0.43             & 0.57             & 0.42          & 0.43         & 0.40          \\ 
\hline
NaiveBayes                   & 0.38             & 0.62             & 0.38          & 0.38         & 0.38         \\ 
\hline
MultilayerPerceptron         & 0.40              & 0.60              & 0.39          & 0.40          & 0.38         \\ 
\hline
SMO                          & 0.43             & 0.57             & 0.42          & 0.43         & 0.42         \\ 
\hline
RandomForest                 & 0.45             & 0.55             & 0.45          & 0.45         & 0.45         \\ 
\hline
\hline
\textbf{Correlation Based Feature Selection}   & \multicolumn{5}{l|}{}                                                             \\ 
\hline
RandomTree                   & 0.54             & 0.46             & 0.54          & 0.54         & 0.54         \\ 
\hline
RandomForest                 & 0.49             & 0.51             & 0.49          & 0.49         & 0.49         \\ 
\hline
\hline
\textbf{Wrapper Subset Eval Feature Selection} & \multicolumn{5}{l|}{}                                                             \\ 
\hline
Random Tree                  & 0.61             & 0.39             & 0.61          & 0.61         & 0.61         \\ 
\hline
Random Forest                & 0.59             & 0.41             & 0.59          & 0.59         & 0.59         \\
\hline
\end{tabular}
 \label{tbl_detection}
	\vspace{-3mm}
\end{table}

Our classification results are presented in Table \ref{tbl_detection}.
Among the tested
algorithms with all of the features chosen, the best \textit{FM} value was found to be 47\% in the RT algorithm. In this algorithm, the correct predicted positive rate
was 47\%, while the incorrect predicted positive rate was 53\%. Further, to
improve the results, we applied some feature selection algorithms. First,
using the correlation-based feature selection algorithm, we conducted the
classification model with the best subset of features
{\cite{hall1999correlation}}. The result indicated the best \textit{FM} as 54\%
again in the RT algorithm. Then, we performed the Wrapper Subset Selection
algorithm and obtained the highest \textit{FM} value of 61\% for the RT algorithm
and 59\% \textit{FM} for the RF algorithm. 

Consequently, these classification results show that identifying the difference
between fake and real news from neural activation is quite difficult even for
machine-learning algorithms.  Therefore, our task performance results and the
accuracy of these classifiers are similar (close to that of random guessing), 
and most standard machine learning techniques do not seem to be able to detect the
differences between the brain signals of fake and reals news trials.

\section{Discussion and Future Directions} 
\label{Discussion}
\subsection{Summary and Insights}
 

The participants in our study showed significantly increased activation in reading news
articles than resting state. The EEG band, Delta, and the EEG metrics, Attention and Meditation, which
are related to decision making, recall, alertness, and consciousness, indicated
higher neural activation when participants were engaged with news articles
than in the resting state. However, the result also shows that participants'
neural activations were not different between the fake and real news articles,
and they could not do a good job of identifying real vs. fake news articles
behaviorally.  We also did a categorical analysis to make sure if the brain
reacts differently in different categories of news articles, including
politics, daily news, and other news. We noticed higher brain activation when
participants were reading articles than the resting state in each
of the categories, but still no differences between real and fake news articles for each category. 
Further, the between-category comparisons of the neural activations
revealed a significant difference in the Attention metric of daily fake news articles
and other fake news articles.

Overall, these results suggest that the users were certainly putting a
considerable effort in reading the news articles of different categories and making real vs. fake decisions,
but they could not make a sound decision in identifying real and fake news
articles as reflected by their brain activity and behavioral responses. 
Although this lack of statistical differences does not necessarily mean that
differences between real news trials and fake news trials do not exist definitively,
our results cast serious doubt as to the presence of such differences.
Perhaps the poor neural and behavioral differences between real vs. fake news
identification was an outcome of the unawareness of the participants as to what
may constitute fake news precisely. Another reason could be that the article
selection was very good for real and fake types, so it may have been hard
to differentiate between the real and the fake articles. Also, we had hidden the actual
URLs in our experiment and therefore there was no chance to
identify the real or fake news without reading the content of the article.
Nevertheless, these findings are unique especially given that subconscious neural activity
differences have been found in the real vs. fake decision-making of other
artifacts such as websites in phishing attacks and paintings while users were
still failing at these tasks behaviorally as reported in prior studies
\cite{huang2011human,neupaneEEG,neupanefNIRS,neupane2014neural}}. 

A consequence of the above finding is that human users may be truly susceptible to
fake news attacks because their brains themselves may not be capable to spot
the differences between fake vs. real news articles. In this light, secure
online social media and news media should not rely upon user inputs or their
brain signals to detect fake news, but rather automated technical approaches
should be emphasized that can alert the users as to the possible presence of
such attacks.  However, our results have a significant real-world implication
to the context of automated detection methods as well: since the AI techniques
for fake news detection often have to crucially rely upon human decision-making
(e.g. for labeling to train and re-train classification models) (e.g.,
\cite{oshikawa2018survey,8966734,wang2017liar,HadeerA2017}), our result
suggests that these techniques themselves may not be as robust as one would
expect in classifying real vs. fake news.
As a case in point, in the study of \cite{MIT}, a manually labeled dataset was
cross-checked, and it was found that some of the true (real) articles are not
true (real) while some false (fake) articles are not false (fake). This
study helps support our argument that human-annotated news may give rise to
biased training models in automated approaches. As a result, we believe that
even machine learning and AI-based fake news detection methods may not be
reliable, in light of our results.

As our other key result, we found significant differences in the Attention metric of daily-other pair of fake news trials. 
This implies that our participants were more engaged in reading other fake news
articles than daily fake news articles. This is perhaps because they may not
have found close similarity in daily fake news articles with real-life
instances but the other fake news articles might have been more interesting or
relevant to them, thus they may have read those with full awareness or
attentiveness. We believe it is an interesting finding that the human brain may
react differently to different types of fake news articles, which may make it
possible to learn what category of fake news would be more interesting to the
users. This could allow fake news designers and other malicious actors to
perform targeted attacks against users based on a specific category of fake
news that users might find more engaging and might be more likely to believe
in.  

As far as improving users' behavioral performance and neural activation in the
real vs. fake news detection task, one possibility is the use of specialized
training programs (e.g., \cite{Interland, Badnews}).  Such training may help users look for specific cues that
may help identify fake news from real news.  The impact of such training
programs on users’ ability to detect fake news could be explored in future
research.

We conducted a posthoc power analysis and calculated observed power values. We see that the observed power of metrics pairs that yielded statistically significant differences are 99\%, 98.7\%, 100\%, and 97.4\%, while others are below 70\%. We believe that our study has enough power since significant pairs have the power of more than 90\%.
\subsection{Study Strengths and Limitations} 

We believe that
our study has several strengths. The neurophysiological sensor was chosen for our
study is a lightweight, easy-to-wear, and wireless EEG headset that allowed us to collect
brain waves almost transparently \cite{neuroskyconf}. 
Another strength is that
we focused on the content of the news articles rather than the side information
such as URLs and advertisements. This is important because, as frequently seen
in phishing attacks, people usually ignore to assess the URLs. Also, fake news
can often be hosted on trustworthy news sites (not just alternative or
malicious sites) for which the users already trust the URLs.

Similar to any other study involving human subjects, our study also had certain
limitations. The study was conducted in a lab environment, which may have
impacted the performance of the participants since they might not have felt the
real security risks. 
Our participants' sample comprised a majority of young students. This
represents a common constraint underlying many university lab studies. However,
our sample exhibited some diversity concerning educational backgrounds.
Moreover, our sample, especially in terms of age, was closer to the group of
users who use the Internet frequently and who are supposedly more vulnerable to
disinformation. Future studies might be needed to further validate our results
with broader participant samples and demographic groups.

One limitation of our study pertains to the number of trials presented to the
participants. Although multiple trials are a norm in EEG (and brain-imaging)
experimental design to achieve a good quality signal-to-noise ratio, the
participants may hardly face many security-related (news trials in our case) in a short span of
time in real life. Our behavioral results, in the fake news experiment, are
still well-aligned with prior work \cite{Vosoughi1146,michaelbarthel}. 

Another limitation related to our machine learning-based detection is deep learning techniques may be able to
identify differences that our standard machine learning techniques could not.
This would need to be explored in future work but would require larger
datasets.

Finally, our participants were explicitly asked to identify a
news article as real or fake. However, in a real-world attack, the victims are
driven to a fake news article from some primary task (e.g., personal social
media account page) and the decision about the legitimacy of the article needs to be made
implicitly. Nevertheless, the users ultimately have to decide the legitimacy of the article. Our results show that, despite
being asked explicitly, users (and their brains) are not able to detect the legitimacy of the
article accurately, and therefore the result may be even worse in a real-world
attack where the decisions are to be made implicitly.
\section{Concluding Remarks} 
\label{CFW}

In this paper, we investigated disinformation, or fake news attack
susceptibility of human users via experimental neuro-cognitive science,
conducting an EEG study of human-centered fake news detection. We identified
the neural underpinnings that control people's responses to real news articles
and fake news articles as they attempt to identify the legitimacy of these articles by reading
through them.  We showed that there are significant differences in neural
activation when users are evaluating real or fake news articles vs. resting,
and when they are evaluating different types of fake news articles.  However,
we did not observe statistically significant, or machine learnable, differences
in neural activities when users were subject to real vs. fake news articles,
irrespective of their behavioral response. We believe that this
neuro-cognitive insight from our work helps justify the users' susceptibility to
fake news attacks as also demonstrated by our behavioral task performance
result.


\bibliographystyle{plain}
\bibliography{sections/main}

\end{document}